\documentclass[12pt]{article}
\usepackage[top=0.51in,bottom=0.51in,left=1.25in,right=1.25in]{geometry}

\usepackage{subfig}
\usepackage{amsmath}
\usepackage{amssymb}
\usepackage{amsthm,lineno}
\usepackage{graphicx}
\usepackage{float}
\usepackage{algorithm}
\usepackage{algorithmic}
\usepackage{multirow}
\usepackage{url}

\def \EC {\mathcal{E}}

\def \RC {\mathcal{R}}
\def \LC {\mathcal{L}}

\def \RB {\mathbb{R}}

\def \hb {\mathbf{h}}
\def \rb {\mathbf{r}}
\def \tb {\mathbf{t}}
\def \wb {\mathbf{w}}
\def \db {\mathbf{d}}

\begin{document}

\title{ParaGraphE: A Library for Parallel Knowledge Graph Embedding}
\author{Xiao-Fan Niu, Wu-Jun Li\\
National Key Laboratory for Novel Software Technology\\
Department of Computer Science and Technology, Nanjing University, China\\
\textit{niuxf@lamda.nju.edu.cn, liwujun@nju.edu.cn}\\
}
\maketitle

\begin{abstract}
Knowledge graph embedding aims at translating the knowledge graph into numerical representations by transforming the entities and relations into continuous low-dimensional vectors. Recently, many methods~\cite{transe, transh, transr, transd, manifolde} have been proposed to deal with this problem, but existing single-thread implementations of them are time-consuming for large-scale knowledge graphs. Here, we design a unified parallel framework to parallelize these methods, which achieves a significant time reduction without influencing the accuracy. We name our framework as ParaGraphE, which provides a library for \underline{para}llel knowledge \underline{graph} \underline{e}mbedding. The source code can be downloaded from~\url{https://github.com/LIBBLE/LIBBLE-MultiThread/tree/master/ParaGraphE}.
\end{abstract}

\section{Introduction}
Knowledge graph is widely used for representing knowledge in artificial intelligence. A knowledge graph contains a set of entities denoted as $\EC$, and a set of relations denoted as $\RC$. Each knowledge in a knowledge graph is represented in a triple form (\emph{head, label, tail}), which is typically denoted as $(h,r,t)$, with $h, t \in \EC$ and $r \in \RC$, meaning that $h$ has a relation $r$ to $t$.

Knowledge graph usually suffers from incompleteness. A main task of knowledge graph research is to complete it by predicting potential relations based on existing observed triples. Traditional methods based on logic and symbols are neither tractable nor robust to deal with large-scale knowledge graphs. Recently, knowledge graph embedding is introduced to solve this problem by encoding entities and relations into continuous low-dimensional vectors, and to perform reasoning in a simple linear algebra way. Some representative knowledge graph embedding methods include TransE~\cite{transe}, TransH~\cite{transh}, TransR~\cite{transr}, TransD~\cite{transd} and ManifoldE~\cite{manifolde}.

Although these methods have achieved promising performance on knowledge graph completion, the existing single-thread implementations of them are time-consuming for large-scale knowledge graphs. Here, we design a unified parallel framework to parallelize these methods, which achieves a significant time reduction without influencing the accuracy. We name our framework as ParaGraphE, which provides a library for \underline{para}llel knowledge \underline{graph} \underline{e}mbedding. The source code can be downloaded from~\url{https://github.com/LIBBLE/LIBBLE-MultiThread/tree/master/ParaGraphE}.

\section{Implemented Methods}
We first introduce the unified framework of our parallel implementations, and then point out the detailed difference between different methods.

\subsection{Framework}
All the methods mentioned above try to minimize a margin-based rank loss which can be formulated as follows:

\begin{align}\label{eq:loss}
\LC &= \sum_{(h,r,t) \in S}\sum_{(h',r,t') \in S'}{[s(h,r,t) + \gamma - s(h', r, t')]_{+}} \\
&s.t. \textit{  some constraints} \nonumber
\end{align}
where $S$ is a set of golden (positive) triples, $S'$ is a set of corrupted (negative) triples which are usually constructed by replacing $h$ or $t$ of a golden triplet $(h,r,t)$ with another entity, $\gamma$ is the margin, $s(h,r,t)$ is the score function of triplet $(h,r,t)$ which is differently defined in different methods.

The unified parallel learning framework of ParaGraphE is shown in Algorithm \ref{alg:paragraphe}.

\begin{algorithm}
\caption{ParaGraphE}
\label{alg:paragraphe}
\begin{algorithmic}
\STATE {Initialization: $p$ threads, embeddings of each entity and relation}
\FOR {$epoch=1$; $epoch<num\_epoches$; $epoch++$}
\FOR {each thread}
\STATE {i=1}
\REPEAT
\STATE {sample a golden triple $(h,r,t)$}
\STATE {construct a corrupted triple $(h',r,t')$}
\IF {$s(h,r,t) + \gamma - s(h',r,t') > 0$}
\STATE {calculate the gradient of the embeddings of $h,t,h',t',r$}
\STATE {subtract the gradient from the corresponding embeddings}
\ENDIF
\STATE normalize $h,t,h',t',r$ according to the constraints
\STATE {i++}
\UNTIL {i== number of triples a thread handles in an epoch}
\ENDFOR
\ENDFOR
\STATE {save the embeddings into files}
\end{algorithmic}
\end{algorithm}

ParaGraphE is implemented based on the lock-free strategies in~\cite{hogwild,DBLP:conf/aaai/ZhaoZL17}.

\subsection{Implemented Methods}
Any method with a similar margin-based rank loss in~(\ref{eq:loss}) can be easily implemented in ParaGraphE. Here, we implement some representative knowledge graph embedding methods include TransE~\cite{transe}, TransH~\cite{transh}, TransR~\cite{transr}, TransD~\cite{transd} and ManifoldE~\cite{manifolde}. All these methods can be formulated with the loss in~(\ref{eq:loss}). The differences between these methods only lie in the score functions and constraints.

\begin{itemize}
	\item TransE~\cite{transe}:
	\begin{itemize}
		\item embeddings: $k$-dimensional vectors for both entities and relations. Let us denote the embeddings for $h,r,t$ as $\hb, \rb, \tb$.
		\item score function: $s(h,r,t)=\Vert \hb + \rb - \tb \Vert$. Either L1-norm or L2-norm can be employed.
		\item constraints: $\Vert \hb \Vert_2 = 1$, $\Vert \tb \Vert_2 = 1$, $\Vert \rb \Vert_2 = 1$.
	\end{itemize}
	\item TransH~\cite{transh}:
	\begin{itemize}
		\item embeddings: $k$-dimensional vectors $\hb, \tb$ for entities $h, t$. A normal vector $\wb_r$ and a translation vector $\db_r$, both in $\RB^k$, are associated with relation $r$.
		\item score function: $s(h,r,t)=\Vert (\hb-{\wb_r}^{\top}\hb\wb_r) + \db_r - (\tb-{\wb_r}^{\top}\tb\wb_r)\Vert$.
		\item constraints: $\Vert \hb \Vert_2 = 1$, $\Vert \tb \Vert_2 = 1$, $\Vert \wb_r \Vert_2 = 1$, $\Vert \db_r \Vert_2 = 1$ and $\vert \wb_r^{\top}\db_r \vert \leq \epsilon$. Here, $\epsilon$ is a hyper-parameter.
	\end{itemize}
	\item TransR~\cite{transr}:
	\begin{itemize}
		\item embeddings: $k$-dimensional vector $\hb, \tb$ for entities $h, t$. A translation vector $\rb \in \RB^d$ and a projection matrix $\mathbf{M}_r \in \RB^{k \times d}$ are associated with relation $r$.
		\item score function: $s(h,r,t)=\Vert \mathbf{M}_r\hb + \rb - \mathbf{M}_r\tb\Vert$.
		\item constraints: $\Vert \hb \Vert_2 = 1$, $\Vert \tb \Vert_2 = 1$, $\Vert \rb \Vert_2 = 1$, $\Vert \mathbf{M}_r\hb \Vert_2 \leq 1$, $\Vert \mathbf{M}_r\tb \Vert_2 \leq 1$.
	\end{itemize}
	\item TransD~\cite{transd}:
	\begin{itemize}
		\item embeddings: Two vectors are associated with each entity and relation, i.e., $\{\hb, \hb_p \in \RB^{k}\}$ for entity $h$, $\{\tb, \tb_p \in \RB^{k}\}$ for entity $t$, and $\{\rb, \rb_p \in \RB^{d}\}$ for relation $r$. Subscript $p$ denotes the projection vectors.
		\item score function: $s(h,r,t)=\Vert (\hb + {\hb_p}^{\top}\hb\rb_p) + \rb - (\tb + {\tb_p}^{\top}\tb\rb_p))\Vert$
		\item constraints: $\Vert \hb \Vert_2 = 1$, $\Vert \tb \Vert_2 = 1$, $\Vert \rb \Vert_2 = 1$, $\Vert \hb + {\hb_p}^{\top}\hb\rb_p \Vert_2 \leq 1$, $\Vert \tb + {\tb_p}^{\top}\tb\rb_p \Vert_2 \leq 1$
	\end{itemize}
	\item SphereE~\cite{manifolde}:
	\begin{itemize}
		\item embeddings: $k$-dimensional vectors for both entities and relations, i.e., $\hb, \rb, \tb \in \RB^{k}$ for $h,r,t$.
		\item score function: $s(h,r,t)=\Vert M(h,r,t) - {D_r}^{2}\Vert$. $M(h,r,t)=\Vert \hb+\rb-\tb\Vert$. $D_r$ is a relation-specific parameter.
		\item constraints: $\Vert \hb \Vert_2 = 1$, $\Vert \tb \Vert_2 = 1$, $\Vert \rb \Vert_2 = 1$.
	\end{itemize}
	
\end{itemize}

Note that SphereE is an implementation for the method ``sphere" of the  ManifoldE~\cite{manifolde}. For more details about these methods, please refer to their original papers.

\section{Experiment Results}

We use 10 threads to run the experiments on two widely used knowledge graph datasets WN18 and FB15k. Table~\ref{table:dataset} lists the statistics of these two datasets. 
 
 \begin{table}[!h]
\centering
\caption{Datasets used in the experiments.}
\label{table:dataset}
\begin{tabular}{l|lllll}
\hline
Dataset & \#Rel & \#Ent & \#Train & \#Valid & \#Test \\ \hline
WN18             & 18             & 40943          & 141442           & 5000             & 5000            \\ \hline
FB15k            & 1345           & 14951          & 483142           & 50000            & 59071           \\ \hline
\end{tabular}
\end{table}

The metrics for evaluation contain those popular metrics adopted in existing knowledge graph embedding papers. They are:
\begin{itemize}
	\item mr: the value of mean rank.
	\item mrr: the value of mean reciprocal rank.
	\item hits@10: the proportion of ranks no larger than 10.
	\item hits@1: the proportion of ranks list at first.
\end{itemize}

The experiment results are shown in Table~\ref{table:wn18} and Table~\ref{table:fb15k}. Here, ``raw" denotes the metrics calculated on all corrupted triples, ``filter" denotes the metrics calculated on corrupted triples without those already existing in knowledge graph.

\begin{table}[!h]
\centering
\caption{Experiment results on WN18.}
\label{table:wn18}
\begin{tabular}{|c|c|c|c|c|c|c|c|c|c|}
\hline
\multirow{2}{*}{method} & \multirow{2}{*}{time(s)} & \multicolumn{4}{c|}{raw}       & \multicolumn{4}{c|}{filter}    \\ \cline{3-10}
                        &                          & mr  & mrr   & hits@1 & hits@10 & mr  & mrr   & hits@1 & hits@10 \\ \hline
TransE                  & 22.6                     & 216 & 0.269 & 0.052  & 0.713   & 204 & 0.346 & 0.080  & 0.830   \\
TransH                  & 18.8                     & 230 & 0.318 & 0.110  & 0.729   & 217 & 0.411 & 0.163  & 0.853   \\
TransR                  & 165.0                    & 224 & 0.396 & 0.202  & 0.774   & 211 & 0.544 & 0.320  & 0.920   \\
TransD                  & 144.2                    & 196 & 0.371 & 0.168  & 0.762   & 181 & 0.510 & 0.274  & 0.909   \\
SphereE                 & 47.1                     & 607 & 0.579 & 0.455  & 0.807   & 593 & 0.871 & 0.819  & 0.941   \\ \hline
\end{tabular}
\end{table}

\begin{table}[!h]
\centering
\caption{Experiment results on FB15k.}
\label{table:fb15k}
\begin{tabular}{|c|c|c|c|c|c|c|c|c|c|}
\hline
\multirow{2}{*}{method} & \multirow{2}{*}{time(s)} & \multicolumn{4}{c|}{raw}       & \multicolumn{4}{c|}{filter}    \\ \cline{3-10}
                        &                          & mr  & mrr   & hits@1 & hits@10 & mr  & mrr   & hits@1 & hits@10 \\ \hline
TransE                  & 61.3                     & 235 & 0.162 & 0.073  & 0.344   & 120 & 0.253 & 0.145  & 0.468   \\
TransH                  & 89.1                     & 222 & 0.188 & 0.086  & 0.405   & 87  & 0.326 & 0.199  & 0.579   \\
TransR                  & 226.7                    & 220 & 0.204 & 0.098  & 0.436   & 75  & 0.378 & 0.242  & 0.640   \\
TransD                  & 149.7                    & 203 & 0.204 & 0.096  & 0.439   & 66  & 0.379 & 0.241  & 0.650   \\
SphereE                 & 147.8                    & 214 & 0.252 & 0.125  & 0.513   & 117 & 0.464 & 0.331  & 0.699   \\ \hline
\end{tabular}
\end{table}

We can find that all the accuracy is very close to the accuracy reported in the original papers, except for SphereE on FB15k where the original paper only reports a result using a polynomial kernel. We can also find that our implementations are much faster than those in the original papers. 

\begin{figure}[!h]
\centering
\subfloat[epoch loss vs time]{\includegraphics[width=0.525\textwidth]{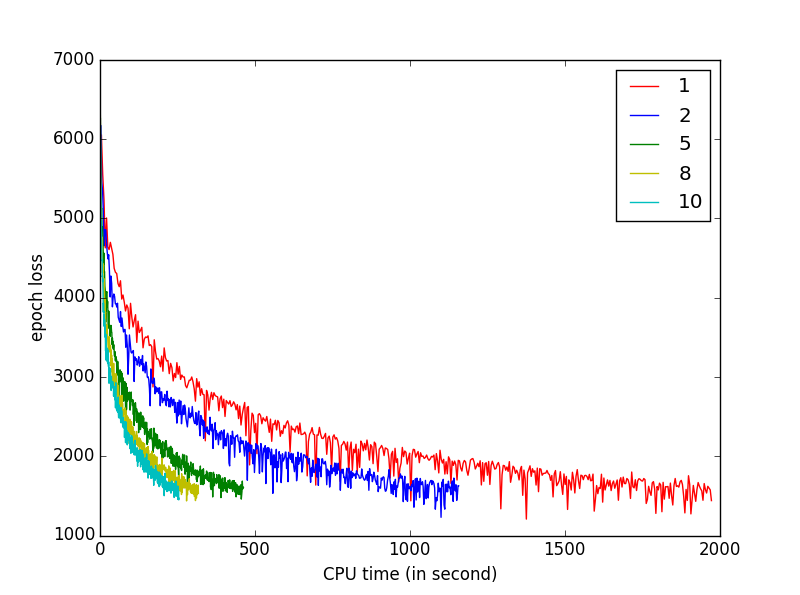}}
\subfloat[epoch loss vs \#epoches]{\includegraphics[width=0.525\textwidth]{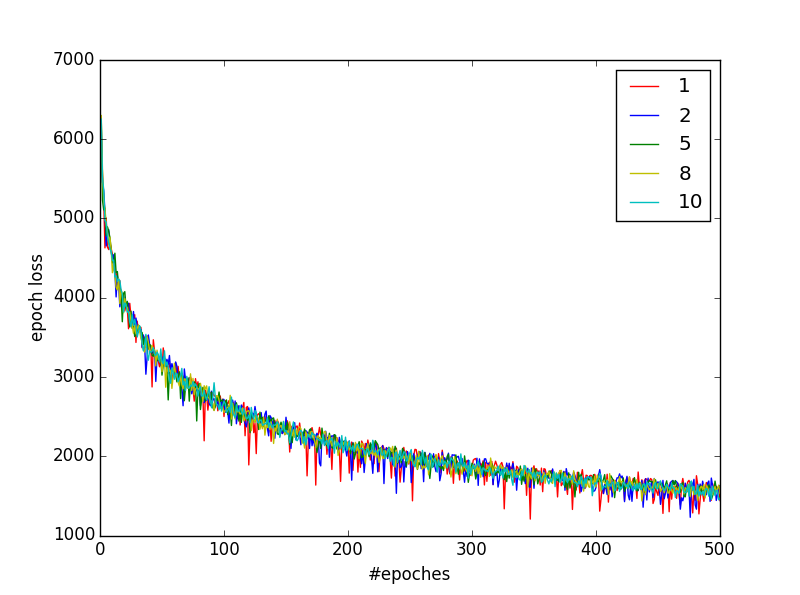}}
\caption{Epoch loss change when training TransR on FB15k.}
\label{fig:loss}
\end{figure}

Figure~\ref{fig:loss} shows the change of epoch loss when running TransR on dataset FB15k with ParaGraphE by setting $\#threads=1, 2, 5, 8, 10$.  Here, the epoch loss is the sum of rank-based hinge loss on the sampled triples in an epoch. We can find that adopting multi-thread learning does not change the convergence of the training procedure. Other embedding methods have similar phenomenon.

\begin{figure}[!h]
\centering
\subfloat[WN18]{\includegraphics[width=0.525\textwidth]{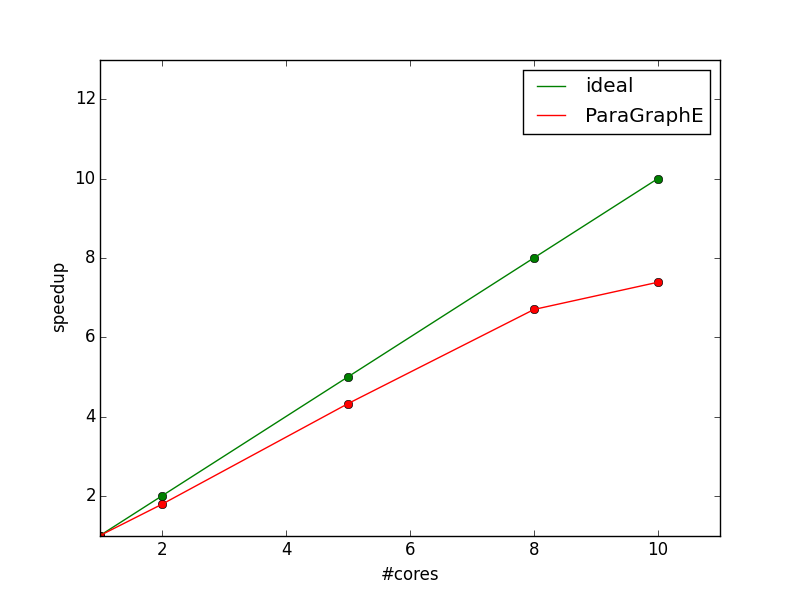}}
\subfloat[FB15k]{\includegraphics[width=0.525\textwidth]{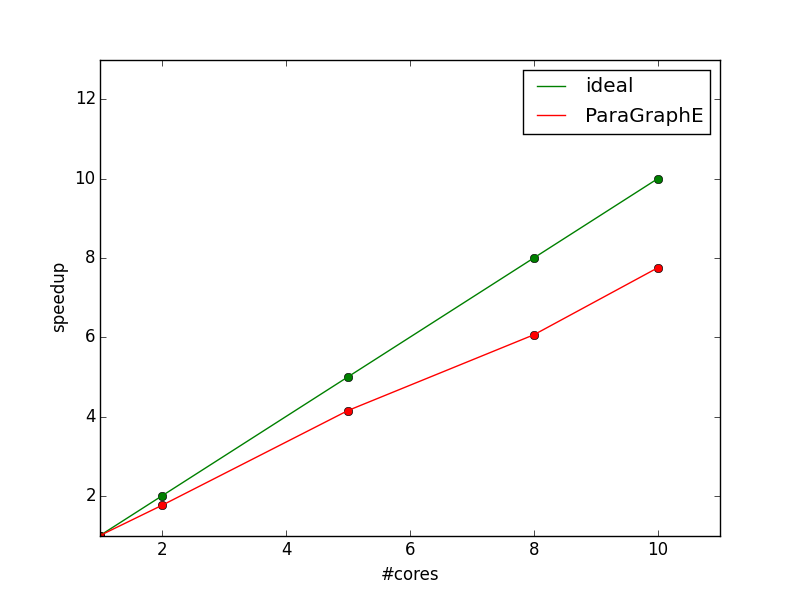}}
\caption{The speedup of training TransR on WN18 and FB15k.}
\label{fig:speed_up}
\end{figure}

Figure~\ref{fig:speed_up} shows the speedup of training TransR on both datasets with ParaGraphE by running 500 epoches. We can find that ParaGraphE achieves around 8x speedup with 10 threads~(cores) on both datasets.

\section{Conclusion}
We have designed a unified framework called ParaGraphE to parallelize knowledge graph embedding methods. Our implementations of several existing methods can achieve a significant time reduction without influencing the accuracy. In our future work, we will implement other knowledge graph embedding methods with the framework ParaGraphE. Moreover, besides knowledge graphs, ParaGraphE is actually general enough to be applied for other kinds of graphs, which will also be empirically evaluated in our future work.

\bibliographystyle{plain}
\bibliography{ParaGraphE}

\end{document}